# The Principle of Least Sensing

## A Privacy-Friendly Sensing Paradigm for Urban Big Data Analytics

*Leye Wang, Peking University, China*

*With the worldwide emergence of data protection regulations, how to conduct law-regulated big data analytics becomes a challenging and fundamental problem. This article introduces the principle of least sensing, a promising sensing paradigm toward law-regulated big data analytics.*

With the prevalence of ubiquitous sensors, our society has witnessed a spectrum of intelligent urban applications in various domains such as smart transportation and planning. Meanwhile, researchers have recognized privacy risks incurred by big data analytics, even on governments' open data [1]. As a result, data protection laws and regulations have become prevalent all around the world, e.g., GDPR (General Data Protection Regulation) of the EU, CCPA (California Consumer Privacy Act) of the US, and PIPL (Personal Information Protection Law) of China. These laws and regulations have pointed out restrictions and guidelines for big data analytics involving personal data. A key character of today's urban big data analytics and smart city applications is *people-centric* [2]. Hence, without a carefully designed process of data sensing from urban residents, urban big data applications can violate laws and regulations. This could further breach user privacy and harm citizen trust in governments and companies. This article articulates a concise guideline, named *the principle of least sensing*, for building a privacy-friendly sensing process for urban big data analytics. Specifically, this principle is proposed to satisfy two widely-accepted data regulation guidelines, i.e., *purpose limitation* and *data minimization*.

**The Principle of Least Sensing: Meeting Purpose Limitation and Data Minimization**

'*Purpose Limitation*' and '*Data Minimization*' are two important points that are explicitly documented in many data regulations and laws. With GDPR as an example, purpose limitation and data minimization have been listed as two principles that data processing parties should follow.[1] More specifically, the purpose limitation principle points out that personal data can only be collected and used for a specified legitimate reason; the data minimization principle demonstrates that the collected personal data should be limited to what is necessary for the purpose. With both purpose limitation and data minimization in mind, to facilitate urban big data analytics, a concise guideline, denoted as the *principle of least sensing*, is articulated as:

***Principle of Least Sensing***: *When conducting urban big data analysis involving personal data, sense and collect only the minimum information necessary for the specified analysis purpose.*

It is worth noting that, if we see the whole city as a ubiquitous computing operation system [3], urban residents' data are precious and sensitive resources of the system. The principle of least sensing is thus a partial analog to the widely known *principle of least privilege* [4], which '*requires that in a particular abstraction layer of a computing environment, every module (such as a process, a user, or a program, depending on the subject) must be able to access only the information and resources that are necessary for its legitimate purpose*'.[2]

---

[1] Purpose Limitation: https://ico.org.uk/for-organisations/guide-to-data-protection/guide-to-the-general-data-protection-regulation-gdpr/principles/purpose-limitation/
Data Minimization: https://ico.org.uk/for-organisations/guide-to-data-protection/guide-to-the-general-data-protection-regulation-gdpr/principles/data-minimisation/

While the principle of least sensing is a high-level guideline for urban data analytics, a gap still needs to be filled for building a practical data sensing process following this principle.

**An Executive Framework of Least Sensing on Data Quantity**

For building a practical data sensing process with the principle of least sensing, the first question is how to define the '*minimum information for the specified analysis purpose*'. More specifically, how to interpret '*minimum*'? The first and perhaps most intuitive explanation of '*minimum*' is on the data quantity, i.e., sensing the smallest amount of data required for the purpose. With this in mind, an illustrative sensing framework is given, denoted as *minimum-quantity sensing*, as shown in Fig. 1.

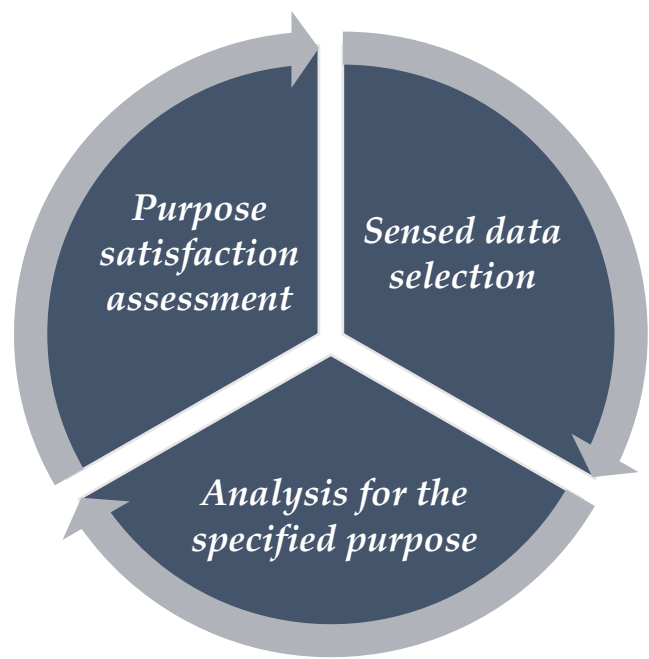

**Figure 1 Minimum-Quantity Sensing Framework**

In general, the minimum-quantity sensing framework is an iterative data sensing and collection process with three main steps: (i) *sensed data selection*, (ii) *analysis for the specified purpose*, and (iii) *purpose satisfaction assessment*. Especially, the three steps constitute a control loop. First, we select the most important data for the specified purpose. Then, with the already collected data, we run the analysis or the computing procedure for the specified purpose. Afterward, we assess whether the currently collected data can satisfy the specified purpose or not. If the answer is 'no', we then continue sensing more data until the specified purpose can be well satisfied.

For a specific urban data analysis application, we need to instantiate the procedures in each step of the minimum-quantity sensing framework. We take an urban big data application, *crowdsensing-*

---
[2] https://en.wikipedia.org/wiki/Principle_of_least_privilege

*based air quality monitoring*, as an example to illustrate details. Crowdsensing is a sensing paradigm that selects people as participants to conduct a target task, e.g., air quality monitoring [5][6]. Specifically, selected participants would use mobile air quality sensors [7] to obtain air quality readings of their stayed places. After the crowdsensing server collects participants' readings, the purpose is to learn an air quality monitoring map of the whole city. While it is inevitable that certain urban regions may not have participants due to human mobility uncertainties, prior research has suggested the inference error of missing-data regions as an effective utility measurement for such a crowdsensing campaign [8]. With these in mind, we can formulate the least-sensing air quality crowdsensing campaign as an optimization process:

$$\min Data\ Quantity \ \ s.t.\ Inference\ Error < \varepsilon$$

Then, following the minimum-quantity sensing framework, a crowdsensing campaign for air quality monitoring can be conducted with three key steps:

- *Participant/Region Selection (sensed data selection)*: Different regions' sensed air quality data can contribute diversely to the overall sensing map. For instance, if sensed data are collected from a limited geospatial range (e.g., only the south part of a city), then it would be hard to accurately infer the air quality of the other urban regions (e.g., the north part of a city). To achieve an optimal selection of the participants and regions to sense, literature has shown that advanced computing techniques, such as active learning and reinforcement learning, can be helpful [9].

- *Whole-city Air Quality Inference (analysis for the specified purpose)*: With collected data from partial regions of a city, then the whole-city air quality sensing map is inferred. Various spatiotemporal inference algorithms can be leveraged here, from traditional statistical learning to recent deep learning methods [10].

- *Inference Error Assessment (purpose satisfaction assessment)*: To ensure that collected data are minimized to satisfy the analysis purpose, determining when data are enough for the purpose is critical. For air quality monitoring, we thus need to ensure that the inference error of non-sensed regions should be small enough (smaller than a predefined threshold). Such an assessment is generally non-trivial as we do not know the real air quality readings of those non-sensed regions. By analyzing the probability distributions of inference errors, literature has shown that statistics tools, such as Bayesian inference, can help assess the inference errors well [11].

For more details about such a three-step control loop for crowdsensing campaigns, interested readers can refer to our published papers [8][11].

**Least Sensing Beyond Data Quantity: Precision, Sensitivity, Predictability, and so on**

Now, let's revisit the potential interpretations of '*minimum*' in the principle of least sensing. We have clarified one interpretation of 'minimum', i.e., data quantity, while in practice the meaning of 'minimum' can be significantly extended. Potential extensions are:

- *Data Precision*. In general, any personal data can be reported with diverse precision granularities. For instance, the location information can be reported as a GPS coordinate, or only as a community area; user ages can be precise up to the birthdate, or just rounding to a 5-year range. The more fine-grained data may incur more

- *Data Sensitivity*. Different types of sensed data may satisfy the same analysis purpose, while these data types could indicate diverse privacy sensitivities for users. For instance, to analyze users' daily life patterns for building a smart home environment, a straightforward solution is deploying cameras at home and then conducting computer vision analytics. However, videos recorded by these cameras are highly sensitive and the leakage of such data can severely harm user privacy [12]. In comparison, recent research achievements have successfully leveraged other low-sensitive data sources, such as WiFi signals, to accurately recognize user behaviors at home [13]. Such a data source change from high sensitivity to low sensitivity can also be regarded as an interpretation of 'minimum' in the least sensing principle.

- *Data Predicability*. In the big data era, state-of-the-art machine learning and data mining techniques can mine deeply hidden correlations between various data. As a result, even if we do not collect privacy-sensitive personal data (e.g., age, residence locations, and political opinions), these data may still be predicted from other data that users probably think 'non-sensitive' and thus make published (e.g. users' public Facebook likes) [14]. Besides reducing the collection of users' sensitive data, a least-sensing system can also aim to minimize the predictability of these sensitive data by controlling and optimizing the collection of other 'non-sensitive' data [15].

In addition to the above interpretations of 'minimum' in the principle of least sensing, there may be more possibilities to explore. Essentially, whatever the interpretation of 'minimum', the design proposal of the least-sensing system should always keep the following optimization formula in mind:

$$\min Data \ \text{s.t.} \ Specified \ Purpose$$

The instantiation of the formula needs careful implementation depending on applications. For instance, in the aforementioned crowdsensing-based air quality monitoring, we want to minimize the data quantity while ensuring the inference error of non-sensed regions lower than a predefined threshold.

**Beyond Sensing: Data Reusing and Sharing by Anonymization**

The principle of least sensing focuses on data sensing *for a specified purpose*. In big data analytics, many important findings are effectively revealed by using data for a purpose *different from the original collection purpose*. It is worth noting that, when we want to reuse or share data for a different purpose, more actions need to be done. One of the most important actions is data *anonymization*. GDPR documents that "*The principles of data protection should therefore **not apply to anonymous information**, namely information which does not relate to an identified or identifiable natural person or to personal data rendered anonymous in such a manner that the data subject is not or no longer identifiable*."[3] That is, anonymous data can be reused and shared for a different purpose. However, data anonymization is not as simple as user ID removal; many users may still be identified from information other than IDs [1][16]. How to do data anonymization is still a challenging task, and interested readers may refer to more materials on this topic [17].

**Conclusion**

This article presents the *principle of least sensing*, which is a high-level guideline to construct a law-regulated and privacy-friendly sensing process for urban big data analytics. The key point is that only minimum data required for a specified analysis purpose should be collected. Moreover, how to interpret 'minimum' is non-trivial; some possible interpretations are given, including data quantity, precision, sensitivity, and predictability. While this article's illustrative examples are mainly on urban big data applications, the principle of least sensing may also inspire other big data application domains that depend highly on personal data, such as healthcare and finance.

---
[3] https://gdpr.eu/recital-26-not-applicable-to-anonymous-data/